\pdfoutput=1 

\documentclass[11pt]{article}

\usepackage{emnlp2021}

\usepackage{times}
\usepackage{latexsym}

\usepackage{hyperref}

\usepackage[T1]{fontenc}

\usepackage[utf8]{inputenc}

\usepackage{amssymb}
\usepackage{microtype}
\usepackage{pifont}%
\usepackage{amsmath,bm,amsfonts,dsfont,bbm}
\usepackage{multirow}
\usepackage{booktabs}

\newcommand\Mark[1]{\textsuperscript#1}

\title{On Generalization in Coreference Resolution}

\author{Shubham Toshniwal\Mark{1}\thanks{~~Equal Contribution}~,
Patrick Xia\Mark{2}\footnotemark[1]~,
Sam Wiseman\Mark{3},
Karen Livescu\Mark{1},
Kevin Gimpel\Mark{1}\\
\Mark{1}Toyota Technological Institute at Chicago\\
\Mark{2}Johns Hopkins University\\
\Mark{3}Duke University\\[0.5em]
\small{\texttt{\{shtoshni, klivescu, kgimpel\}@ttic.edu,
paxia@cs.jhu.edu,
swiseman@cs.duke.edu}}\\}

\begin{document}
\maketitle
\begin{abstract}
While coreference resolution is defined independently of dataset domain, most models for performing coreference resolution do not transfer well to unseen domains.
We consolidate a set of 8 coreference resolution datasets targeting different domains to evaluate the off-the-shelf performance of models. 
We then mix three datasets for training; even though their domain, annotation guidelines, and metadata differ, we propose a method for jointly training a single model on this heterogeneous data mixture by using data augmentation to account for annotation differences and sampling to balance the data quantities. 
We find that in a zero-shot setting, models trained on a single dataset transfer poorly while joint training yields improved overall performance, leading to better generalization in coreference resolution models. 
This work contributes a new benchmark for robust coreference resolution and multiple new state-of-the-art results.\footnote{Code available at \url{https://github.com/shtoshni92/fast-coref}}
\end{abstract}

\section{Introduction}
Coreference resolution is a core component of the NLP pipeline, as determining which mentions in text refer to the same entity is used %
for a wide variety of downstream tasks like knowledge extraction \cite{li-etal-2020-gaia}, question answering \cite{dhingra-etal-2018-neural}, and dialog systems \cite{gao-etal-2019-interconnected}. 
As these tasks span many domains, we need coreference models to generalize well. 

Meanwhile, 
models for coreference resolution have improved due to neural architectures with millions of parameters and the emergence of pretrained %
encoders. However, model generalization across domains has always been a challenge %
\cite{yang-etal-2012-domain, zhao-ng-2014-domain, poot-van-cranenburgh-2020-benchmark, aktas-etal-2020-adapting}. Since these models are %
usually engineered for a single dataset, they capture idiosyncrasies inherent in that dataset. %
As an example, OntoNotes \cite{weischedel2013ontonotes}, a widely-used general-purpose dataset, provides metadata, like the document genre and speaker information. However, this assumption cannot be made more broadly, especially if the input is raw text~\cite{wiseman-etal-2016-antecedent}. %

\begin{table*}[ht]
    \centering
    \small
        \begin{tabular}{lccrccccc}
    \toprule
         & \multicolumn{3}{c}{Num. Docs} & Words/ &  Mentions/ &Mention & Cluster & \% of singleton \\
         Dataset & Train & Dev. & Test  & Doc &Doc &  length &  size & mentions\\
    \midrule
         OntoNotes  & \phantom{1}2802 & 343 & 348 & \phantom{1}467 & \phantom{1}56 & 2.3 & 4.4 &  \phantom{1}0.0 \\
         LitBank\textsuperscript{k} & \phantom{11}80 & \phantom{1}10 & 10 & 2105 & 291 & 2.0 & 3.7 & 19.8 \\
         PreCo & 36120 & 500 & 500  & \phantom{1}337 & 105 & 2.7 & 1.6 & 52.0 \\
         Character Identification & \phantom{11}987 & 122 & 192  & \phantom{1}262 & \phantom{1}36 & 1.0 & 5.1 &  \phantom{1}6.4 \\
          WikiCoref & \phantom{1111}0 & \phantom{11}0 & 30 & 1996 & 230 & 2.6 & 5.0 & \phantom{1}0.0 \\
         Quiz Bowl Coreference\textsuperscript{k} & \phantom{1111}0 & \phantom{11}0 & 400 & \phantom{1}126 & \phantom{1}24 & 2.7 & 2.0 & 26.0 \\
         Gendered Ambiguous Pronouns \textsuperscript{p} & \phantom{1}2000 & 400 & 2000 &  \phantom{11}95 & \phantom{11}3 & 2.0 & - & - \\
         Winograd Schema Challenge\textsuperscript{p} & \phantom{1111}0 & \phantom{11}0 & 271 & \phantom{11}16 & \phantom{11}3 & 1.5 & - & -  \\
    \bottomrule
    \end{tabular}
    \caption{Statistics of datasets. Datasets with \textsuperscript{k} indicate that prior work uses \textit{k}-fold cross-validation; we record the splits used in this work. Datasets with \textsuperscript{p} are partially annotated, so we do not include cluster details.}
    
    \label{tab:data}
\end{table*}

Furthermore, while there are datasets aimed at capturing a broad set of genres \cite{weischedel2013ontonotes, poesio-etal-2018-anaphora, zhu-etal-2021-ontogum}, they are not mutually compatible due to differences in annotation guidelines. For example, some datasets do not annotate singleton clusters (clusters with a single mention). %
Ideally, we would like a coreference model to be robust to all the standard datasets. 
In this work, we consolidate 8 datasets spanning multiple domains, document lengths, and annotation guidelines. We use them to evaluate the off-the-shelf performance of models trained on a single dataset. While they perform well %
within-domain (e.g., a new state-of-the-art of 79.3 F1 on LitBank), they still perform poorly out-of-domain. 

To address poor out-of-domain performance, we propose joint training for coreference resolution, which is challenging due to the incompatible training procedues for different datasets.  Among other things, we need to address (unannotated) singleton clusters, as OntoNotes does not include singleton annotations. We propose a data augmentation process to add predicted singletons, or \textit{pseudo-singletons}, into the training data to match the other datasets which have gold singleton annotations. %

Concretely, we contribute a benchmark for coreference to highlight the disparity in model performance and track generalization. We find joint training highly effective and show that including more datasets is almost ``free'', as performance on any single dataset is only minimally affected by joint training. %
We find that our data augmentation method of adding pseudo-singletons 
is also effective. With all of these extensions, we increase the macro average F\textsubscript{1} across all datasets by 9.5 points and achieve a new state-of-the-art on LitBank and WikiCoref.

\section{Datasets}
\label{sec:datasets}

We organize our datasets into three types. %
Training datasets (Sec.~\ref{sec:data:train}) are large in terms of number of tokens and clusters and more suitable for training. Evaluation datasets (Sec.~\ref{sec:data:eval}) are out-of-domain compared to our training sets and are entirely held out. Analysis datasets (Sec.~\ref{sec:data:an}) contain annotations aimed at probing specific phenomena. \autoref{tab:data} lists the full statistics.

\subsection{Training Datasets}
\label{sec:data:train}

\paragraph{OntoNotes 5.0 (ON)} \cite{weischedel2013ontonotes} is a collection of news-like, web, and religious texts spanning seven distinct genres. Some genres are transcripts (phone conversations and news). As the primary training and evaluation set for developing coreference resolution models, many features specific to this corpus are tightly integrated into publicly released models. For example, the metadata includes information on the document genre and the speaker of every token (for spoken transcripts). Notably, it does not contain singleton annotations. %

\paragraph{LitBank (LB)} \cite{bamman-etal-2020-annotated} is a set of public domain works of literature drawn from Project Gutenberg. On average, coreference in the first 2,000 tokens of each work is fully annotated for six entity types.\footnote{They are people, facilities, locations, geopolitical entities, organizations, and vehicles.} %
We only use the first cross-validation fold of LitBank, which we call LB\textsubscript{0}.

\paragraph{PreCo (PC)} \cite{chen-etal-2018-preco} contains documents from reading comprehension examinations, each fully annotated for coreference resolution. Notably, the corpus is the largest such dataset released. %

\subsection{Evaluation Datasets}
\label{sec:data:eval}

\paragraph{Character Identification (CI)} \cite{zhou-choi-2018-exist} has multiparty conversations derived from TV show transcripts. Each scene in an episode is considered a separate document. 
This character-centric dataset only annotates mentions of people. %

\paragraph{WikiCoref (WC)} \cite{ghaddar-langlais-2016-wikicoref} contains documents from English Wikipedia. This corpus contains sampled and annotated documents of different lengths, from 209 to 9,869 tokens. 

\paragraph{Quiz Bowl Coreference (QBC)} \cite{guha-etal-2015-removing} contains questions from Quiz Bowl, a trivia competition. These paragraph-long questions are dense with entities. Only certain entity types (titles, authors, characters, and answers) are annotated.

\begin{table*}[t]
    \centering
    \small
    \begin{tabular}{llccccccccc}%
    \toprule
          Model & Training & ON & LB\textsubscript{0} & PC & CI & WC & QBC & GAP & WSC & Macro Avg.\\
   \midrule
        longdoc &  ON 
            & 79.0   & 54.8    & 44.3  & 49.8 & 59.6      & 36.8     & 88.9 & 59.8                                    & 59.1 \\
        longdoc \textsuperscript{S} &  ON 
            & 79.6 & 54.6    & 44.0    & 58.7 & 60.1      & 36.4     & \textbf{89.8} & 59.4                                    & 60.3    \\
        longdoc \textsuperscript{S, G} &  ON 
           & 79.5 & 54.7 & 44.5 & 59.5 & 59.9 & 37.0 & 89.0 & 58.7 & 60.3 \\
        longdoc \textsuperscript{S} &  ON + PS 60K 
            & \textbf{80.6} & 56.6    & 49.1  & 55.6 & 62.1      & 40.1     & 89.3 & 61.3                                    & \textbf{61.8}\\
        longdoc &  LB\textsubscript{0}
             & 56.6 & 77.2    & 46.8  & 53.3 & 47.5      & \textbf{50.5}     & 85.3 & 32.8 & 56.3\\
        longdoc &  PC
            & 58.8 & 50.3    & \textbf{87.8}  & 39.5 & 50.7      & 46.5     & 87.3 & \textbf{62.7} & 60.5  \\
    \midrule
        longdoc \textsuperscript{S} & Joint
           & 79.2 & \textbf{78.2}    & 87.6  & 59.4 & 60.3      & 42.9     & 88.6 & 60.1 & 69.5 \\
        longdoc \textsuperscript{S} & Joint + PS 30K      & 79.6 & \textbf{78.2}  & 87.5  & 58.4 & \textbf{62.5}      & 45.5     & 88.7 & 59.4 & \textbf{70.0} \\
    \bottomrule
    \end{tabular}
    \caption{Performance of each model on 8 datasets measured by CoNLL F\textsubscript{1} \cite{pradhan-etal-2012-conll}, except for GAP (F\textsubscript{1}) and WSC (accuracy). Some models use speaker (\textsuperscript{S}) features, genre (\textsuperscript{G}) features, or pseudo-singletons (PS).} %
    \label{tab:results}
\end{table*}

\subsection{Analysis Datasets}
\label{sec:data:an}

\paragraph{Gendered Ambiguous Pronouns (GAP)} \cite{webster-etal-2018-mind} is a corpus of ambiguous pronoun-name pairs derived from Wikipedia. While only pronoun-name pairs are annotated, they are provided alongside their full-document context. This corpus has been previously used to study gender bias in coreference resolution systems. %

\paragraph{Winograd Schema Challenge (WSC)} \cite{levasque2012wino} is a challenge dataset for measuring common sense in AI systems.\footnote{\url{https://cs.nyu.edu/~davise/papers/WinogradSchemas/WSCollection.html}} 
Unlike the other datasets, each document contains one or two sentences with a multiple-choice question.
We manually align the multiple choices to the text and remove 2 of the 273 examples due to plurals.
\section{Models}

\subsection{Baselines}

We first evaluate a recent system \cite{xu-choi-2020-revealing} which extends a mention-ranking model \cite{lee-etal-2018-higher} by making modifications in the decoding step. We find disappointing out-of-domain performance and difficulties with longer documents present in LB\textsubscript{0} and WC (Appendix \ref{sec:appendix:other:xu}). To overcome this issue, we study the \emph{longdoc} model by \citet{toshniwal-etal-2020-learning}, which is an entity-ranking model designed for long documents that reported strong results on both OntoNotes and LitBank. %

The original longdoc model uses a pretrained SpanBERT~\cite{joshi-etal-2020-spanbert} encoder which we replace with Longformer-large~\cite{Beltagy2020Longformer} as it can incorporate longer context.
We retrain the longdoc model and finetune the Longformer encoder for each dataset, which proves to be competitive for coreference.\footnote{The model scores 79.5 on OntoNotes and achieves state-of-the-art on LitBank with 79.3. Details are in Appendix \ref{sec:appendix:other:lb}.} %
For OntoNotes we train with and without the metadata of: (a) genre embedding, and (b) speaker identity which is introduced as part of the text as in~\citet{wu-etal-2020-corefqa}. %

\subsection{Joint Training}

With copious amounts of text in OntoNotes, PreCo, and LitBank, we can train a joint model on the combined dataset. However, this is impractical as the annotation guidelines between the datasets are misaligned (OntoNotes does not annotate singletons and uses metadata) and because there are substantially more documents in PreCo. 

\paragraph{Augmenting Singletons} Since OntoNotes does not annotate for singletons, our training objective for OntoNotes is different from that of PreCo and LitBank. To address this, we introduce \textit{pseudo-singletons} that are \textit{silver} mentions derived from first training a mention detector on OntoNotes and selecting the top-scoring mentions outside the gold mentions.\footnote{This mention detector is architecturally the first half of the longdoc model.} We experiment with adding 30K, 60K, and 90K 
pseudo-singletons (in total, there are 156K gold mentions). We find  adding 60K to be the best fit for OntoNotes-only training, and 30K is the best for joint training (Appendix \ref{sec:singleton_ontonotes}).   

\paragraph{Data Imbalance}
PreCo has 36K training documents, compared to 2.8K and 80 training documents for OntoNotes and 
LitBank respectively. A naive dataset-agnostic sampling strategy would mostly sample PreCo documents. To address this issue, we downsample OntoNotes and PreCo to 1K documents per epoch. Downsampling to 0.5K documents per epoch led to slightly worse performance (Appendix \ref{sec:downsampling joint}).

\paragraph{Metadata Embeddings}
For the joint model to be applicable to unknown domains, we avoid using any domain or dataset-identity embeddings, including the OntoNotes genre embedding.  
We do make use of speaker identity in the joint model because: (a) this is possible to obtain in conversational and dialog data, and (b) it does not affect other datasets that are known to be single-speaker at test time. %

\begin{table*}
		\setlength{\tabcolsep}{0pt}

	\begin{tabular}{p{0.1\textwidth}p{0.9\textwidth}}
	\toprule
	Dataset & Instance \\
	\midrule
	(1) QBC & (\colorbox{blue!30}{\textbf{This poem}}) is often considered \colorbox{blue!30}{the counterpart of another poem} \dots name \colorbox{blue!30}{this poem about a creature ``burning bright, in the forests of the night,"} \dots\\\midrule
	(2) QBC & This author's non fiction works \dots \colorbox{blue!30}{another work}, a plague strikes secluded valley where teenage boys have been evacuated \dots name this author of \colorbox{blue!30}{Nip the Buds, Shoot the Kids}
	\dots\\\midrule

	(3) QBC & This poet of ``(\textbf{I}) felt a Funeral in (\textbf{my}) Brain" and ``I'm Nobody, Who are you?"  wrote about a speaker who hears a Blue, uncertain, stumbling buzz before expiring in ``(\textbf{I}) heard a fly buzz when (\textbf{I}) died". For 10 points, name this female American poet of Because (\textbf{I}) could not stop for Death.\\\midrule
		(4) CI & \textit{Chandler Bing:} Okay, I don't sound like that. (\textbf{That}) is so not true. (\textbf{That}) is so not ... (\textbf{That}) is so not ... That ... Oh , shut up !\\
	\bottomrule
	\end{tabular}
\caption{Joint + PS 30K error analysis for zero-shot evaluation sets. Each row highlights one cluster where spans in parenthesis are predicted by the model while the blue-colored spans represent ground truth annotations. Thus, in (2) the model misses out on the ground truth cluster entirely while in  (3) and (4) the model predicts an additional cluster.
}

\label{tab:error}
\end{table*}

\section{Results}

Table~\ref{tab:results} shows the results for all our models on all 8 datasets. We report each dataset's associated metric (e.g., CoNLL F\textsubscript{1}) and a macro average across all eight datasets to compare generalizability.

Among the longdoc baseline models trained on one of OntoNotes, PreCo, or LitBank, we observe a sharp drop in out-of-domain evaluations. 
The LitBank model is generally substantially worse than the models trained on OntoNotes and PreCo, likely due to both a smaller training set and a larger domain shift. Interestingly, the LitBank model performs the best among all models on QBC, which can be attributed to both LB and QBC being restricted to a similar set of markable entity types. 
Meanwhile, all OntoNotes-only models perform well on WC and GAP, possibly due to the more diverse set of genres within ON and because WC also does not contain singletons.

For models trained on OntoNotes, we find that the addition of speaker tokens leads to an almost 9 point increase on CI, which is a conversational dataset, but has little impact for non-conversational evaluations.
Surprisingly, the addition of genre embeddings has almost no impact on the overall evaluation.\footnote{In fact, we find that for the model trained with genre embeddings, modifying the genre value during inference has almost no impact on the final performance.}
Finally, the addition of pseudo-singletons leads to consistent  significant gains across almost all the evaluations, including OntoNotes.

The joint models, which are trained on a combination of OntoNotes, LitBank, and PreCo, suffer only a small drop in performance on OntoNotes and PreCo, and achieve the best performance for LitBank. 
Like the results observed when training with only OntoNotes, we see a significant performance gain with pseudo-singletons in joint training as well, which justifies our intuition that they can bridge the annotation gap.
The ``Joint + PS 30K'' model also achieves the state of the art for WC.

\section{Analysis}
\paragraph{Impact of Singletons}

Singletons are known to artificially boost the coreference metrics~\cite{kubler-zhekova-2011-singletons}, and their utility for downstream applications is arguable. 
To determine the impact of singletons on final scores, we present separate results for singleton and non-singleton clusters in QBC in Table~\ref{tab:singleton_ment}. 
For non-singleton clusters we use the standard CoNLL F\textsubscript{1} but for singleton clusters the CoNLL score is undefined, and hence, we use the vanilla F\textsubscript{1}-score. 

The poor performance of ON models for singletons is expected, as singletons are not seen during training. Adding pseudo-singletons improves the performance of both the ON and the Joint model for singletons. Interestingly, adding pseudo-singletons also leads to a small improvement for non-singleton clusters. 

The PC model has the best performance for non-singleton clusters while the LB\textsubscript{0} model, which performs the best in the overall evaluation, has the worst performance for non-singleton clusters. This means that the gains for the LB\textsubscript{0} model can be all but attributed to the superior mention detection performance which can be explained by the fact that both LB and QBC are restricted to a similar set of markable entity types. 

\paragraph{Impact of Domain Shift}
Table~\ref{tab:error} presents instances where the Joint + PS 30K model makes mistakes. 
In examples (1) and (2), the model misses out on mentions referring to literary works which is because references to literary texts are rare in the joint training data. 
Example (2) also requires world knowledge to make a connection between the description of the work and its title. 
In example (3) the model introduces an extraneous cluster consisting of first person pronouns mentioned in titles of different works. The model lacks the domain knowledge that narrators across different works are not necessarily related.  
Apart from the language shift, there are annotation differences across datasets as well. For example (4) drawn from CI, the model predicts a valid cluster (for Chandler Bing's speaking style) according to the ON annotation guidelines but the CI dataset doesn't annotate such clusters.

\begin{table}[t!]
	\centering
	\small
	\begin{tabular}{lccc}
		\toprule
		Data & Singleton & Non-singleton & Overall \\
		\midrule
		ON &  \phantom{1}0.4 & 43.9 & 36.4 \\
		ON + PS 60K &  14.4 & 44.4 & 40.1 \\
		LB\textsubscript{0} & \textbf{44.9} & 41.2 & \textbf{50.5} \\
		PC & 28.8 & \textbf{50.3} & 46.5 \\
		Joint & 21.7 & 47.3 & 42.9 \\
		Joint + PS 30K  & 26.7 & 48.6 & 45.5 \\ 
		\bottomrule
	\end{tabular}
	\caption{Performance on singleton and non-singleton clusters for QBC. ON=longdoc\textsuperscript{S} and PS=pseudo-singletons.}
	\label{tab:singleton_ment}
\end{table}
\section{Related work}
Joint training is commonly used 
in NLP for training robust models,
usually aided by learning dataset, language, or domain embeddings
(e.g.,~\cite{stymne-etal-2018-parser} for parsing;~\cite{kobus-etal-2017-domain, tan-etal-2019-multilingual} for machine translation). %
This is essentially what models for OntoNotes already do with genre embeddings \cite{lee-etal-2017-end}. Unlike prior work, our test domains are 
unseen, so we cannot 
learn test-domain embeddings. 

For coreference resolution, \citet{aralikatte-etal-2019-rewarding} %
augment annotations using relation extraction systems to better incorporate world knowledge, a step towards generalization. \citet{subramanian-roth-2019-improving} use adversarial training to target names, %
with improvements on GAP. 
\citet{moosavi-strube-2018-using} incorporate linguistic features to improve generalization to WC.
Recently, \citet{zhu-etal-2021-ontogum} proposed the OntoGUM dataset which consists of multiple genres. However, compared to the datasets used in our work, OntoGUM is much smaller, and is also restricted to a single annotation scheme.
To the best of our knowledge, our work is the first to evaluate generalization at scale. 

Missing singletons in OntoNotes has been previously addressed through new data annotations, leading to the creation of the ARRAU \cite{poesio-etal-2018-anaphora} and PreCo \cite{chen-etal-2018-preco} corpora. While we include PreCo in this work, ARRAU contains additional challenges, like split-antecedents, that further increase the heterogeneity, and its domain overlaps with OntoNotes. Pipeline models for coreference resolution 
that first detect mentions naturally leave behind unclustered mentions as singletons, although understanding singletons can also 
improve performance \cite{recasens-etal-2013-life}. 

Recent end-to-end neural models have been evaluated on OntoNotes, and therefore conflate ``not a mention'' with ``is a singleton'' \cite{lee-etal-2017-end, lee-etal-2018-higher, kantor-globerson-2019-coreference, wu-etal-2020-corefqa}.
For datasets with singletons, this has been addressed explicitly through a cluster-based model \cite{toshniwal-etal-2020-learning, yu-etal-2020-cluster}. For those without, they can be implicitly accounted for with auxiliary objectives \cite{zhang-etal-2018-neural, swayamdipta-etal-2018-syntactic}. We go one step further 
by augmenting with \textit{pseudo-singletons}, so that the training objective is identical regardless of whether the training set contains annotated singletons. %

\section{Conclusion}
Our eight-dataset benchmark highlights disparities in coreference resolution model performance and tracks cross-domain generalization.
Our work begins to address cross-domain gaps, first by handling differences in singleton annotation via data augmentation with pseudo-singletons, and second by training a single model jointly on multiple datasets.  This approach produces promising improvements in generalization, as well as new state-of-the-art results on multiple datasets.  We hope that future work will continue to use this benchmark to measure progress towards truly general-purpose coreference resolution.

\section*{Acknowledgements}
This material is based upon work supported by the National Science Foundation under Award No.~1941178.

\clearpage

\bibliographystyle{acl_natbib}
\bibliography{main} 

\clearpage
\newpage %

\appendix

\section{Model and Training Details}

\subsection{Model}

Our model follows the typical coreference pipeline of encoding the document, followed by mention proposal, and  finally mention clustering. The model is architecturally the same as \citet{toshniwal-etal-2020-learning}, and so we re-present their formulation throughout this section. However, we use the Longformer encoder as it accommodates longer documents. 
Otherwise, the model is identical to \citet{toshniwal-etal-2020-learning} in terms of model size and weight dimensions. We next explain the mention proposal and mention clustering stages briefly.

\paragraph{Mention Proposal}
Given a document $\mathcal{D}$, we score all mentions of length $ \leq 20$ subword tokens and choose the $K = 0.4 \times |\mathcal{D}|$ top spans among them. This is an initial pruning step that speeds up the model and reduces memory usage. Let $X(K) = \{(x_i)_{i=1}^K\}$ represent the top-$K$ candidate mention spans and $s_m(x_i)$ be a learned scoring function for span $x_i$, which represents how likely a span is an entity mention. 
$s_m$ is trained to assign positive score to gold mentions (any mention in a gold cluster), and negative score otherwise. The training objective only uses spans in $X(K)$, i.e. loss is computed after pruning.
During inference, we can therefore further prune down to $\{x_i : x_i \in X(K), s_m(x_i) \geq 0\}$, which we then pass into the clustering step.
During training, we use teacher forcing and only pass gold mentions among the top-$K$ mentions to the clustering step.

\paragraph{Mention Clustering}
The entity-based model tracks $M$ entities (initially $M=0$). Let $E = (e_m)_{m=1}^M$ represent the $M$ entities.
For ease of notation, we will overload the terms $x_i$ and $e_j$ to also correspond to their respective representations.

The model decides whether the span $x_i$ refers to any of the entities in $E$ as follows:
\begin{align*}
s_c(x_i, e_j) &\!=\! f_c([x_i; e_j; x_i \odot e_j; g(x_i, e_j)])\\
s_c^{\mathit{top}} &\!=\! \max_{j=1 \dotsc M} s_c(x_i, e_j)\\
e^{\mathit{top}} &\!=\! \underset{{j=1 \dotsc M}}{\arg\max}\; s_c(x_i, e_j)
\end{align*} %
where $\odot$ represents the element-wise product, and $f_c(\cdot)$ corresponds to a learned feedforward neural network. The term $g(x_i, e_j)$ corresponds to a concatenation of feature embeddings that includes embeddings for (a) number of mentions in $e_j$, and (b) number of tokens between $x_i$ and last mention of $e_j$. If $s_c^{\mathit{top}} > 0$ then $x_i$ is considered to refer to $e^{\mathit{top}}$, and $e^{\mathit{top}}$ is updated accordingly.\footnote{We use weighted averaging where the weight for $e^{\mathit{top}}$ corresponds to the number of previous mentions seen for  $e^{\mathit{top}}$.}
Otherwise, we initiate a new cluster: $E = E \cup x_i$.
During training, we use teacher-forcing i.e. the clustering decisions are based on ground truth.

\subsection{Training}
We train all the models for 100K gradient steps with a batch size of 1 document. Only the LB-only models are trained for 8K gradient steps which corresponds to 100 epochs for LB. The models are evaluated a total of 20 times (every 5K training steps) for all models except the LB-only models which are evaluated every 400 steps. 
We use early stopping and a patience of 5 i.e.\ training stops if the validation performance doesn't improve for 5 consecutive evaluations. 

We use the full context size of 4096 tokens for Longformer-large. All training documents used in this work except 1 ON document fit in a single context window. For optimizer, we use AdamW with a weight decay of 0.01 and initial learning rate of 1e-5 for the Longformer encoder, and Adam with an initial rate of 3e-4 for the rest of the model parameters. The learning rate is linearly decayed throughout the training.

\begin{table*}[t]
    \centering
    \small
    \begin{tabular}{cccccccc}
    \toprule
        \multicolumn{3}{c}{Training} & \multicolumn{5}{c}{Evaluation} \\
        Num ON & Num PC & Num PS & ON & LB & PC & WC & QB\\
    \midrule
        \phantom{1}500 & \phantom{1}500 & \phantom{11}0 & 79.4 & 78.8 & 85.0 & 60.8 & 45.3 \\
        \phantom{1}500 & \phantom{1}500 & 30K  & 79.4 & 79.5 & 84.8 & 61.2 & 47.7 \\
        1000 & 1000 & \phantom{1}0 & 79.7 & 79.4 & 85.1 & 60.3 & 42.9 \\
        1000 & 1000 & 30K & 79.5 & 78.7 & 85.1 & 62.5 & 45.5 \\

    \midrule
    1000 & 1000 & 60K & 78.9 & 77.4 & 85.1 & 61.3 & 46.6 \\
    1000 & 1000 & 90K & 78.5 & 77.7 & 85.1 & 60.7 & 47.4 \\
    \bottomrule
    
    \end{tabular}
    \caption{Validation set scores of datasets when downsampling OntoNotes (ON) and PreCo (PC) in joint training.}
    \label{tab:downsample}
\end{table*}

\begin{table*}[t]
    \centering
    \small
    \begin{tabular}{llccccccccc}%
    \toprule
          & Training & ON & LB & PC & CI & WC & QBC & Macro Avg.\\
    \midrule
    longdoc &  ON & 88.8 & 62.8 & 42.6 & 65.7 & 72.1 & 53.1 & 64.2 \\
    longdoc \textsuperscript{S} &  ON  & 89.2 & 61.7 & 41.8 & 75.5 & 69.9 & 52.5 & 65.1 \\
    longdoc \textsuperscript{S, G} &  ON  & 89.4 & 62.8 & 42.7 & 77.0 & 72.5 & 52.9 & 66.2 \\
    longdoc \textsuperscript{S} &  ON + PS 60K  & 87.7 & 81.5 & 81.0 & 76.1 & 71.6 & 70.4 & 78.0 \\ 
    longdoc &  LB\textsubscript{0} & 77.6 & 85.8 & 81.4 & 67.6 & 69.8 & 70.2 & 75.4 \\
    longdoc &  PC & 76.8 & 81.1 & 93.6 & 66.5 & 67.0 & 77.3 & 77.0 \\
    \midrule
    longdoc \textsuperscript{S} & Joint & 90.9 & 86.9 & 93.4 & 77.2 & 76.6 & 68.7 & 82.3 \\
    longdoc \textsuperscript{S} & Joint + PS 30K  & 89.4 & 87.1 & 93.3 & 77.3 & 74.6 & 71.7 & 82.2 \\

    \bottomrule
    \end{tabular}
    \caption{Result of all the models with gold mentions.
     Some models use speaker (\textsuperscript{s}) features, genre (\textsuperscript{g}) features, or pseudo-singletons (PS).
    The metric for the training and evaluation datasets is CoNLL F-score. 
    We skip the analysis datasets because they lack the set of true gold mentions. 
    }
    \label{tab:results_gold_mentions}
\end{table*}

\section{Other Results}
\label{sec:appendix:other}

\subsection{\citet{xu-choi-2020-revealing} Baselines}
\label{sec:appendix:other:xu}

We run the off-the-shelf model on the test sets of ON, LB\textsubscript{0}, PC, and QBC. LB\textsubscript{0} requires a 24GB GPU, while WC runs out of memory even on that hardware. The model shows strong in-domain performance with 80.2 on ON. However, out-of-domain performance is weak: 57.2 on LB\textsubscript{0}, 49.3 on PC, and 37.6 on QBC. These are roughly on par with the ON longdoc models. 

\subsection{LitBank Cross-Validation Results}
\label{sec:appendix:other:lb}

Table~\ref{table:litbank_all} presents the results for all the cross-validation splits of LitBank. The overall performance of 79.3 CoNLL F1 is state of the art for LitBank, outperforming the previous state of the art of 76.5 by~\citet{toshniwal-etal-2020-learning}. Note that in this work, the joint model outperformed (78.2 vs. 77.2) this baseline model on split 0 (LB\textsubscript{0}). However, training 10 joint models contradicts the purpose of this work, which is to create a single, generalizable model. Realistically, we recommend jointly training with the entirely of LitBank.

\begin{table}[t]
    \centering
    \small
    \begin{tabular}{ccc}%
    \toprule
Cross-val split & Dev & Test \\
\midrule
0 & 78.8 & 77.2 \\
1 & 78.6 & 80.3 \\
2 & 81.1 & 78.7 \\
3 & 79.0 & 79.1 \\
4 & 80.0 & 78.7 \\
5 & 79.7 & 78.7 \\
6 & 77.7 & 80.7 \\
7 & 81.9 & 79.1 \\
8 & 78.0 & 80.8 \\
9 & 81.8 & 78.7 \\
\midrule
Total & 79.7 & 79.3 \\
\bottomrule
\end{tabular}
\caption{LitBank cross-validation results.}
\label{table:litbank_all}
\end{table}

\subsection{Singleton Results for OntoNotes}
\label{sec:singleton_ontonotes}

\begin{table}[t]
    \centering
    \small
    \begin{tabular}{ccc}%
    \toprule
    Num PS & Val. & Test \\
    \midrule
    \phantom{11}0 & 79.9 & 79.6 \\
    30K & 79.9  & 80.5 \\
    60K & \textbf{80.0} & \textbf{80.6} \\
    90K & 79.9 & 80.0 \\
    \bottomrule
    \end{tabular}
    \caption{Validation and Test results for ON-only model trained on different amount of pseduo-singletons (PS). }
    \label{tab:ontonotes_singletons}
\end{table}

For ON-only models, we tune over the number of pseudo-singletons sampled among \{30K, 60K, 90K\}. Table~\ref{tab:ontonotes_singletons} shows that 60K pseudo-singletons is the best choice based on validation set results on ON.

\subsection{Downsampling and Singleton Results for Joint}
\label{sec:downsampling joint}

In preliminary experiments, we sample 500 docs from ON and PC. \autoref{tab:downsample} shows the results, confirming that 1K is slightly better than 500. Using more examples (e.g. 5K PC) begins to hurt performance on LB, likely due to data imbalance. 

For the 1K downsampling setting, we tune over the number of pseudo-singletons sampled among \{30K, 60K, 90K\}. We find 30K to be the best choice based on validation set results.

\subsection{Results with Gold Mentions}
\label{sec:gold_mentions}

In Table~\ref{tab:results_gold_mentions}, we report the results with gold mentions for the training and evaluation sets. The analysis sets are skipped as they are partially annotated. We find that joint training is also helpful in this setting, as results mirror findings with predicted mentions. In particular, this shows that it is not just a failure to predict mentions that is preventing ON from performing well on LB, PC, and QBC.

\section{Compute Resources}
Given that we are finetuning the Longformer model and using a maximum context size of 4096 tokens, the memory requirements of the model are quite large even though the cluster-ranking paradigm is considered memory efficient~\cite{xia-etal-2020-incremental}. 
We were able to train the PreCo-only model on a 12 GB GPU in 20 hrs  (even the longest PreCo documents are shorter than 2048 tokens with the Longformer tokenization). All other models were trained over GPUs with memory 24GB or higher (Titan RTX and A6000). On an A6000, the LB-only models can be trained within 4 hrs, the ON-only models within 16 hrs, and the joint models within 20 hrs.

\end{document}